\documentclass[conference]{IEEEtran}
\IEEEoverridecommandlockouts
\usepackage[ruled,lined,linesnumbered]{algorithm2e}
\usepackage{float}
\usepackage{amsmath, bbm}
\usepackage{amssymb}
\usepackage{amsfonts}
\usepackage{amsthm}
\usepackage{amsbsy}
\usepackage{nicefrac}
\usepackage{comment}
\usepackage{bm}
\usepackage{hyperref}
\usepackage{cite}
\usepackage[utf8]{inputenc} 
\usepackage[T1]{fontenc}
\usepackage{graphicx}
\usepackage{epstopdf}
\usepackage[font=small]{caption}
\usepackage{tikz}
\usetikzlibrary{positioning}
\usepackage{pgfplots}
\usepackage{subcaption}
\usepackage{multirow}
\def\cast{{
   \mathord{
      \hbox to 0em{
         \ooalign{
	   \smash{\hbox{$\ast$}}\crcr
	   \smash{\hskip-1pt\Large\hbox{$\circ$}} }
	 \hidewidth}
      \phantom{\bigcirc}
} }}

\def\bm#1{\mbox{\boldmath $#1$}}

\newcommand{\bds}{\begin {itemize}}
\newcommand{\eds}{\end {itemize}}
\newcommand{\bdf}{\begin{definition}}
\newcommand{\blm}{\begin{lemma}}
\newcommand{\edf}{\end{definition}}
\newcommand{\elm}{\end{lemma}}
\newcommand{\bthm}{\begin{theorem}}
\newcommand{\ethm}{\end{theorem}}
\newcommand{\bprp}{\begin{prop}}
\newcommand{\eprp}{\end{prop}}
\newcommand{\bcl}{\begin{claim}}
\newcommand{\ecl}{\end{claim}}
\newcommand{\bcr}{\begin{coro}}
\newcommand{\ecr}{\end{coro}}
\newcommand{\bquest}{\begin{question}}
\newcommand{\equest}{\end{question}}

\newcommand{\larrow}{{\larrow}}

\newcommand{\argmin}{\ensuremath{\mathrm{arg}\min}}
\newcommand{\argmax}{\ensuremath{\mathrm{arg}\max}}

\newcommand{\cS}{{\ensuremath{\mathcal{S}}}}

\newcommand{\va}{{\ensuremath{{\mathbf{a}}}}}

\newcommand{\vb}{{\ensuremath{{\mathbf{b}}}}}

\newcommand{\vx}{{\ensuremath{{\mathbf{x}}}}}

\newcommand{\vy}{{\ensuremath{{\mathbf{y}}}}}

\newcommand{\mA}{{\ensuremath{\mathbf{A}}}}

\newcommand{\mD}{{\ensuremath{\mathbf{D}}}}

\newcommand{\mL}{{\ensuremath{\mathbf{L}}}}

\newcommand{\mP}{{\ensuremath{\mathbf{P}}}}

\newcommand{\mS}{{\ensuremath{\mathbf{S}}}}

\newcommand{\mU}{{\ensuremath{\mathbf{U}}}}

\newcommand{\mX}{{\ensuremath{\mathbf{X}}}}
\newcommand{\mY}{{\ensuremath{\mathbf{Y}}}}
\newcommand{\mZ}{{\ensuremath{\mathbf{Z}}}}

\usepackage{latexsym}

\def\IC{\mathbb C}
\def\IN{\mathbb N}
\def\IZ{\mathbb Z}
\def\IR{\mathbb R}

\def\shat{^{\mathchoice{}{}%
 {\,\,\smash{\hbox{\lower4pt\hbox{$\widehat{\null}$}}}}%
 {\,\smash{\hbox{\lower3pt\hbox{$\hat{\null}$}}}}}}

\def\bSigma{{
      \ooalign{
      \smash{\hskip.4pt\raise.4pt\hbox{$\Sigma$}}\vphantom{}\crcr
      \smash{\hskip.7pt\raise.6pt\hbox{$\Sigma$}}\vphantom{}\crcr
      \smash{\hbox{$\Sigma$}}\vphantom{$\Sigma$}}
      \vphantom{\hbox{$\Sigma$}}
      }}
\def\bTheta{{
      \ooalign{
      \smash{\hskip.5pt\raise.5pt\hbox{$\Theta$}}\vphantom{}\crcr
      \smash{\hskip.0pt\raise.1pt\hbox{$\Theta$}}\vphantom{}\crcr
      \smash{\hbox{$\Theta$}}\vphantom{$\Theta$}}
      \vphantom{\hbox{$\Theta$}}
      }}
\def\bDelta{{
      \ooalign{
      \smash{\hskip.4pt\raise.4pt\hbox{$\Delta$}}\vphantom{}\crcr
      \smash{\hskip.7pt\raise.6pt\hbox{$\Delta$}}\vphantom{}\crcr
      \smash{\hbox{$\Delta$}}\vphantom{$\Delta$}}
      \vphantom{\hbox{$\Delta$}}
      }}
\def\bLambda{{
      \ooalign{
      \smash{\hskip.5pt\raise.5pt\hbox{$\Lambda$}}\vphantom{}\crcr
      \smash{\hskip.0pt\raise.1pt\hbox{$\Lambda$}}\vphantom{}\crcr
      \smash{\hbox{$\Lambda$}}\vphantom{$\Lambda$}}
      \vphantom{\hbox{$\Lambda$}}
      }}

\makeatletter

\def\bordermatrix#1{\begingroup \m@th
  \@tempdima 8.75\p@
  \setbox\z@\vbox{%
    \def\cr{\crcr\noalign{\kern2\p@\global\let\cr\endline}}%
    \ialign{$##$\hfil\kern2\p@\kern\@tempdima&\thinspace\hfil$##$\hfil
      &&\quad\hfil$##$\hfil\crcr
      \omit\strut\hfil\crcr\noalign{\kern-\baselineskip}%
      #1\crcr\omit\strut\cr}}%
  \setbox\tw@\vbox{\unvcopy\z@\global\setbox\@ne\lastbox}%
  \setbox\tw@\hbox{\unhbox\@ne\unskip\global\setbox\@ne\lastbox}%
  \setbox\tw@\hbox{$\kern\wd\@ne\kern-\@tempdima\left[\kern-\wd\@ne
    \global\setbox\@ne\vbox{\box\@ne\kern2\p@}%
    \vcenter{\kern-\ht\@ne\unvbox\z@\kern-\baselineskip}\,\right]$}%
  \null\;\vbox{\kern\ht\@ne\box\tw@}\endgroup}
\makeatother

\makeatletter
\def\argmin{\mathop{\operator@font arg\,min}}
\def\argmax{\mathop{\operator@font arg\,max}}
\makeatother

\def\bm#1{\mbox{\boldmath $#1$}}

\newcommand{\bbeta}{{\bm\beta}}

\newcommand{\bea}{\begin{array}}
\newcommand{\ena}{\end{array}}
\newcommand{\beq}{\begin{equation}}
\newcommand{\enq}{\end{equation}}

\newcommand{\beqa}{\begin{eqnarray}}
\newcommand{\enqa}{\end{eqnarray}}

\newcommand{\beqan}{\begin{eqnarray*}}
\newcommand{\enqan}{\end{eqnarray*}}

\newcommand{\AL}{\begin{enumerate}}
\newcommand{\ALE}{\end{enumerate}}

\def\addots{\mathinner{
    \mkern1mu\raise0pt\vbox{\kern7pt\hbox{.}}
    \mkern2mu\raise4pt\hbox{.}
    \mkern2mu\raise7pt\hbox{.}
    \mkern1mu}}

\def\sddots{\mathinner{
    \mkern.8mu\raise7pt\hbox{.}
    \mkern.8mu\raise4pt\hbox{.}
    \mkern.8mu\raise0pt\vbox{\kern7pt\hbox{.}}
    \mkern1mu}}

\def\saddots{\mathinner{
    \mkern.2mu\raise0pt\vbox{\kern7pt\hbox{.}}
    \mkern.2mu\raise4pt\hbox{.}
    \mkern.2mu\raise7pt\hbox{.}
    \mkern1mu}}

\def\sqplus{\mathbin{
	{\ooalign{\hfil\raise.3ex\hbox{\scriptsize
	+}\hfil\crcr\mathhexbox274\crcr\mathhexbox275}}
	}} 
\def\sqminus{\mathbin{
	{\ooalign{\hfil\raise.3ex\hbox{\scriptsize
	--}\hfil\crcr\mathhexbox274\crcr\mathhexbox275}}
	}}

\def\IC{{
   \mathord{
      \hbox to 0em{
	 \hskip-4pt
         \ooalign{
	   \smash{\hskip1.9pt\raise2.6pt\hbox{$\scriptscriptstyle |$}}\crcr
	   \smash{\hbox{\rm\sf C}} }
	 \hidewidth}
      \phantom{\hbox{\rm\sf C}}
} }}
\def\IN{
    {\ooalign{
   \smash{\hskip2.2pt\raise1.5pt\hbox{$\scriptscriptstyle |$}}\vphantom{}\crcr
   \hbox{\sf N}
	}}
	} 
\def\IZ{
    {\ooalign{
   \smash{\hskip1.9pt\raise0pt\hbox{$\sf Z$}}\vphantom{}\crcr
   \hbox{\sf Z}
	}}
	} 
\def\IR{
    {\ooalign{
   \smash{\hskip2.2pt\raise1.5pt\hbox{$\scriptscriptstyle |$}}\vphantom{}\crcr
   \smash{\hskip2.2pt\raise3.3pt\hbox{$\scriptscriptstyle |$}}\vphantom{}\crcr
   \hbox{\sf R}
	}}
	} 

\DeclareMathAlphabet{\mathcmb}{OT1}{cmr}{b}{n}

\def\bSigma{\ensuremath{\mathcmb{\Sigma}}}
\def\bLambda{\ensuremath{\mathcmb{\Lambda}}}

\def\bTheta{\ensuremath{\mathcmb{\Theta}}}

\newcommand{\SI}{\begin{indlist}}
\newcommand{\EI}{\end{indlist}}

\newcommand{\DL}{\begin{dashlist}}
\newcommand{\DLE}{\end{dashlist}}

\makeatletter
\def\setboxz@h{\setbox\z@\hbox}
\def\wdz@{\wd\z@}
\def\boxz@{\box\z@}
\def\underset#1#2{\binrel@{#2}%
  \binrel@@{\mathop{\kern\z@#2}\limits_{#1}}}
\def\binrel@#1{\begingroup
  \setboxz@h{\thinmuskip0mu
    \medmuskip\m@ne mu\thickmuskip\@ne mu
    \setbox\tw@\hbox{$#1\m@th$}\kern-\wd\tw@
    ${}#1{}\m@th$}%
  \edef\@tempa{\endgroup\let\noexpand\binrel@@
    \ifdim\wdz@<\z@ \mathbin
    \else\ifdim\wdz@>\z@ \mathrel
    \else \relax\fi\fi}%
  \@tempa
}
\let\binrel@@\relax%
\makeatother

\newcommand\abs[1]{\left|#1\right|}

\AppendGraphicsExtensions{.tif}
\pdfoutput=1
\begin{document}
\title{ Lightweight Industrial Cohorted Federated Learning for Heterogeneous Assets 
}

\author{\IEEEauthorblockN{Madapu Amarlingam\dag, Abhishek Wani\dag\dag, Adarsh NL\dag\dag, }
\IEEEauthorblockA{madapu.amarlingam@in.abb.com, abhishek.aw@iiits.in, adarsh.nl@iiits.in}
\IEEEauthorblockA{ABB Corporate Research Center, Bengaluru, India\dag \\
Department of CSE, IIIT Sricity, Chittor, India\dag\dag
}

}

\maketitle

\begin{abstract}
Federated Learning (FL) is the most widely adopted collaborative learning approach for training Decentralized Machine Learning (ML) models by exchanging learning between clients without sharing the data and compromising privacy. However, since great data similarity or homogeneity is taken for granted in all FL tasks, FL is still not specifically designed for the industrial setting. Rarely this is the case in industrial data because there are differences in machine type, firmware version, operational conditions, environmental factors, and hence, data distribution. Albeit its popularity, it has been observed that FL performance degrades if the clients have heterogeneous data distributions. Therefore, we propose a Lightweight Industrial Cohorted FL ($\texttt{LICFL}$) algorithm that uses model parameters for cohorting without any additional on-edge (client-level) computations and communications than standard FL and mitigates the shortcomings from data heterogeneity in industrial applications. Our approach enhances client-level model performance by allowing them to collaborate with similar clients and train more specialized or personalized models. Also, we propose an adaptive aggregation algorithm that extends the $\texttt{LICFL}$ to Adaptive $\texttt{LICFL}$ ($\texttt{ALICFL}$) for further improving the global model performance and speeding up the convergence. Through numerical experiments on real-time data, we demonstrate the efficacy of the proposed algorithms and compare the performance with existing approaches.   
\end{abstract}

\title{
}

\section{Introduction}

Process industries such as manufacturing, oil, gas, and mining are moving towards Industry 4.0, where edge analytics is the key technology for its success as it enables real-time responses, cost reductions, and ensures the security of data and systems \cite{a1,a2}. The most common applications of edge analytics are predictive maintenance, automated quality control, precision control, and automated guided vehicles \cite{a3}, to name a few.

In process automation, to reduce the wrong predictions/decisions and avoid interruptions of the process plant operations, it is crucial to train a generalized and robust ML model for edge analytics \cite{a4,b6a}. The performance of the ML models directly depends on the volume of the data used for training, which is often only available to a limited degree for individual sensor types at the edge in most industrial applications \cite{a5}. Data sharing within the industry or with an external industry partner can be used to meet the requirement of the data for individual types of sensors. The initial method necessitates a substantial communication bandwidth to transmit complete data from the edges to off-premises servers and is often constrained by resources within the core network infrastructure of process industries \cite{a5a}. Also, the latter approach is not feasible as sensitive information about process parameters or vulnerable information might be disclosed.

\begin{figure}[h!]
	\centering
		\includegraphics[width=0.4\textwidth]{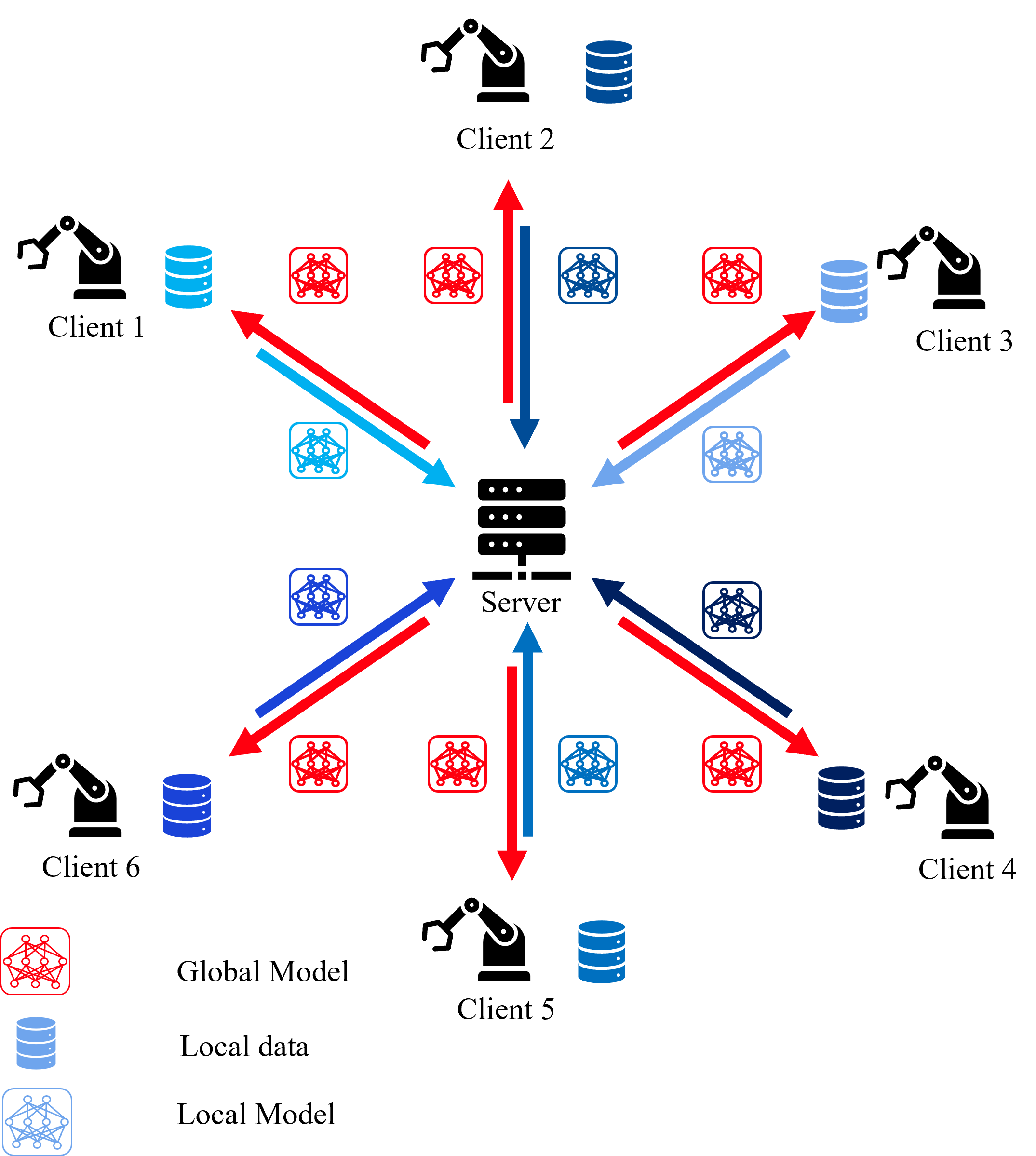}
        \caption{Federated Learning, each client updates the model parameters with local data and sends them to the server, which aggregates and shares them back.}
	\label{fig:FL_new}
\end{figure}

To tackle this issue, a widely used collaborative learning approach Federated Learning (FL) can be used in industrial settings \cite{a6}. FL enables collaborative training of ML models across multiple edge devices without sharing the data \cite{a7}. This approach consists of a central authority called a server, and local edge devices called clients. Each client trains an ML model on their local dataset and then sends their model parameters to the server. The server aggregates the model parameters received from multiple clients and shares the aggregated model (global model) parameters with the clients. The same process is presented in Fig.~\ref{fig:FL_new}. The complete cycle, involving training of local model at client devices, transmission of locally-trained models to the server, aggregation of these models at the server to obtain a global model, and the subsequent dissemination of the global model back to client devices, is collectively termed as one communication round in FL.

\begin{figure}[h!]
	\centering
		\includegraphics[width=0.5\textwidth]
        {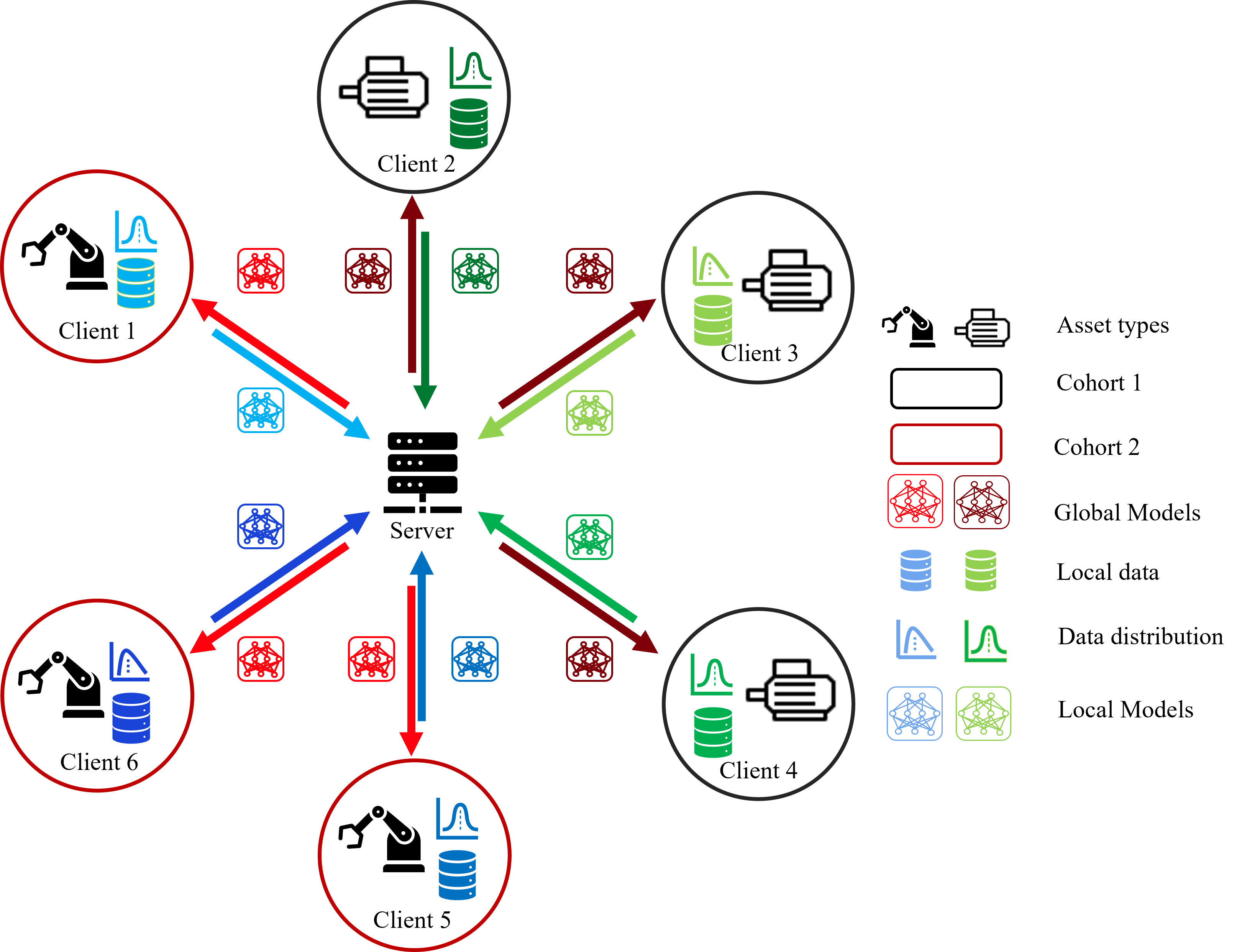}
        \caption{Industrial Federated Learning, machine type is considered as a filter or parameter for cohorting to avoid divergence of the global model and improve performance at the client-level model.}
	\label{fig:IFL_new}
\end{figure}

However, FL is still not specifically made for the industrial environment because high data similarity or data homogeneity is assumed in most of the existing FL algorithms \cite{a7, a8, b123}. Due to changes in the type of machine, firmware version, operational settings, and environmental factors, this is rarely the case in industrial sensor data. Fig.~\ref{fig:IFL_new}, shows a typical scenario of an industrial environment, which depicts two machine types with differences in data distributions. In Fig.~\ref{fig:IFL_new}, machine type is considered as a parameter or filter to cluster (cohort) and they are grouped to avoid heterogeneity in the models \cite{a9}. Despite its widespread use, it has been found that FL diverges if the clients' data distributions are heterogeneous. Recently, in \cite{a10} authors proposed an FL method that performs two-level clustering: in the primary level, the meta-information such as machine type, and firmware version are used for clustering (as shown in Fig.~\ref{fig:IFL_new}) followed by second-level where the individual client computes statistical moments of the data and shares with the server, further the server uses these moments of the data as filters or parameters for second-level clustering (cohorting). Each cluster or cohort contributes to a model, that helps to avoid client-level ML model divergence and poor performance. However, the first four moments of the data may not capture the complete characteristics \cite{a10}. Moreover, computing the moments and sending them to the server entails additional computational overhead and communication costs to the clients. 

In this paper, we propose a novel lightweight cohorted federated learning algorithm, which combines similar clients into groups and conducts FL within each group. This approach mitigates adverse effects arising from heterogeneity in the data distributions across the clients. Our algorithm uses model parameters for cohorting the clients, this makes our approach lightweight and does not require any additional on-edge computations and communications at the client for cohorting. Our method enhances client-level model performance by allowing them to collaborate with similar clients and train more specialized or personalized models. The lightweight cohorted federated learning algorithm augments the primary-level cohorting, and applies it as second-level cohorting. Our contributions are threefold:
\begin{itemize}
    \item Propose a Lightweight Industrial Cohorted Federated Learning ($\texttt{LICFL}$) algorithm that incorporates a plug-and-play model parameter-based cohorting approach, which is agnostic to data, aggregation strategy, and requires no additional on-edge computations and communications at the clients for cohorting.
    \item Propose an algorithm for auto selection of aggregation strategies for each cohort, which extends the $\texttt{LICFL}$ to Adaptive $\texttt{LICFL}$ ($\texttt{ALICFL}$) for improving the global model convergence and accuracy.
    \item Extensive numerical experiments to demonstrate the performance of the proposed algorithms.
\end{itemize}

The remaining portion of the paper is organized as follows: Section II introduces the notations, gives a primer for vanilla FL, and describes the proposed algorithms. Section III presents the considered data set and experimentation methodology. Section IV demonstrates the experimental results with a discussion. Section V concludes the paper. 

\section{Adaptive Lightweight Cohorted Industrial FL}

\subsection{Federated Learning: A review}
Consider a setup with $K$ industrial plants (clients) with a centralized server or cloud, as shown in Fig.~\ref{fig:FL_new}. Suppose the training data available at the $k^{th}$ client is $\{(\vx_s,\vy_s), \forall s \in \cS_k\}$. Generally, data at different plants are mutually disjoint and non-empty sets $\cS_k$ such that the overall data from all the plants can be written as $\cS = \bigcup_{k=1}^K \cS_k$ and $\cS_k \cap \cS_j = \emptyset$ for $k \neq j$. Each plant trains local model parameters at the $k^{th}$ client by minimizing the loss function $\mathcal{L}_{k}({\boldsymbol \Theta}) = \frac{1}{|\cS_k|} \sum_{s \in \cS_{k}} l_{s}(\vy_s,f(\vx_s;{\boldsymbol \Theta}))$, where ${\boldsymbol \Theta}$ represents model parameters. The gradients computed by the clients based on their local data can be written as 
$
\nabla \mathcal{L}_{k} ({\boldsymbol \Theta}) = \frac{1}{|\cS_k|}\sum_{s \in \cS_k}\nabla l_{s}(\vy_s,f(\vx_s;\boldsymbol{\Theta}))$, where $f(\vx_s,\boldsymbol{\Theta}) = \hat{\vy}_{s}$, predicted output.

As the data sets are disjoint, the learning at each plant is limited to corresponding local data. To train a collaborative robust model from disjoint data sets of different plants, FL computes the gradient of $l(\boldsymbol{\Theta})$ with respect to $\boldsymbol{\Theta}$ as
\begin{align*}
\nabla l(\boldsymbol{\Theta}) = \frac{1}{|\cS|} \sum_{s \in \cS} \nabla l_{s}(\vy_s,f(\vx_s,\boldsymbol{\Theta})) &=  \sum_{k=1}^K \frac{|\cS_k|}{|\cS|} \nabla \mathcal{L}_{k}(\boldsymbol{\Theta}), \\
\end{align*}
where $\nabla \mathcal{L}_{k}(\boldsymbol{\Theta})$ denotes the gradient computed at the $k^{th}$ client based on the local data.
In reality, the gradients from the clients can be aggregated at the server to compute the global consensus gradient $\nabla l(\boldsymbol{\Theta})$. Further, the global update of $\boldsymbol{\Theta}$ can be computed by aggregating the local updates from the clients as
\begin{eqnarray}
\text{Local training:}& \quad \quad {\boldsymbol \Theta}_{k} &\gets \quad {\boldsymbol \Theta}_k-\eta_{l} \nabla \mathcal{L}_{k}({\boldsymbol \Theta})  \label{eq1}\\
\text{Aggregation }\operatorname{A} \text{:}& \quad \quad {\boldsymbol \Theta} &\gets \quad \sum_{k=1}^{K} \frac{|\cS_k|}{|\cS|} {\boldsymbol \Theta}_k. \label{eq2}
\end{eqnarray}
In other words, each client computes the current model parameters with its local data, and the server computes a weighted average of all these model parameters and sends it back to the clients. Note that lowercase letters or symbols with bold denote the vector and suffix refers to the element in that vector. Similarly, uppercase letters or symbols with bold denote the matrix and suffixes refer to the elements (row and column) inside the matrix. The same notation is followed everywhere. 

As mentioned in the previous section, in an industrial setting due to heterogeneity in the type of machinery, firmware version, operational settings, environmental factors, and data distribution, the learning of the global consensus model diverges. To accommodate the heterogeneity and provide model personalization, the following subsection proposes Lightweight Industrial Cohorted FL ($\texttt{LICFL})$, which incorporates a model parameter-based cohorting algorithm.  

\subsection{Lightweight Industrial Cohorted FL ($\texttt{LICFL})$}
In general, a neural network model weights are trained as a function of input data, where the weights (or model parameters) capture the data properties \cite{b31}. For example, from equation \eqref{eq1}, one can understand that the updated model parameters at $k^{th}$ client ${\boldsymbol \Theta}_{k}$ are a function of gradient $\mathcal{L}_{k}({\boldsymbol \Theta})$ as it depends on it. The $\mathcal{L}_{k}({\boldsymbol \Theta})$ = $\frac{1}{|\cS_k|} \sum_{s \in \cS_{k}} l_{s}(\vy_s,f(\vx_s;{\boldsymbol \Theta}))$ is a function of input data $\{(\vx_{s},\vy_{s})\}$. Therefore the weights of a neural network ${\boldsymbol \Theta}$ are indeed a function of the data $\{(\vx_{s},\vy_{s})\}$ and capture the properties. By leveraging this property of model parameters, Algorithm~\ref{alg:cohort} proposes a lightweight model parameter-based cohorting methodology that considers only the model weights or parameters (which inherently capture the properties of the data) of all the clients for cohorting and does not require any additional computations or communications at the client.

As described in Algorithm~\ref{alg:cohort}, in the first communication round $(r=1)$, the server sends the initial model ${\boldsymbol \Theta}$ to clients (line: $6$). Clients update the initial model with respect to the loss function $f(\boldsymbol{\Theta})$ and local data $S_{k}$ (line: $30$ to $32$). The $k^{th}$ client sends updated model parameters $\boldsymbol{\Theta}_k$ to the server (line: $33$). When the server receives the set of updated model parameters denoted by $\mathcal{V}$ from all the clients (line: $8$), it executes the model parameter-based cohorting algorithm (line: $10$). The pseudo-code of the parameter-based cohorting algorithm is presented in Algorithm~\ref{alg:ch}.

\subsubsection{Model parameter-based cohorting}

In general, neural network models consist of multiple layers. Without loss of generality, we flatten all the model parameters of $k^{th}$ client $\mathcal{V}_{k}=\boldsymbol{\Theta}_{k}$ into a single vector $\vx_{k}$ for further processing, the same is described in line: $3$ of Algorithm~\ref{alg:ch}. The matrix $\mX$ is formed by concatenating all the model parameter vectors from $K$ clients. Usually higher dimension clustering produces inaccurate clusters \cite{b1a}. To overcome this, inspired by Principle Component Analysis (PCA) \cite{b1}, the Algorithm~\ref{alg:ch} transforms the model parameters $\mX$ to a lower dimension $\mY$ by preserving significant features of the model parameters with high variance, the same is described from line: $5$ to $8$. For accommodating the complex relations among model parameters of the clients, inspired by spectral clustering \cite{b28, b29, b30}, we use graph-based clustering. A weighted graph Adjacency matrix $\mA$ is build from the rows of the matrix $\mY$ (line: $9$) and compute Laplacian $\mL$ (line: $11$). The largest $q$ Eigenvalues are denoted as $\vb$ and corresponding Eigenvectors are denoted as $\mS \in \mathbb{R}^{K\times q}$ (line: $13$), which are computed by Eigenvalue decomposition of the Laplacian matrix $\mL$ (line: $12$). Further, the matrix $\mS$ is normalized ($\mP \in \mathbb{R}^{K\times q}$) and used for clustering (line:$14$), where each row represents a client and column represents data, transformed weights belongs to a client in the new space. For our implementation, we used k-means clustering for grouping the columns of the $\mP$ matrix. The Algorithm~\ref{alg:ch} returns a set of cohorts which is represented by $\mathbf{C}$ (line: $15$).

\begin{algorithm}[h]
    \KwIn{$K$: Number of Clients, $R$: \# Communication Rounds, $\operatorname{A}$: Aggregation Strategy}
    Initialize: $ {\boldsymbol \Theta}, {\boldsymbol C} $ \\
    \SetAlgoLined
    \For{$r=1\dots R$ }{
        \If{ $r = 1$}{ 
            \textbf{Server Executes:} \\
            \For{$k=1 \dots K$}{
                ${\boldsymbol \Theta}_{k} \longleftarrow \texttt{ClientUpdates}({\boldsymbol \Theta},k)$ 
            }
            $  {\mathcal{V}} \longleftarrow \{{\boldsymbol \Theta}_1, {\boldsymbol \Theta}_2 \dots         {\boldsymbol \Theta}_k\} $ \\
            ${\boldsymbol \Theta} \leftarrow \operatorname{A}(\mathcal{V})$\\
            ${\boldsymbol C} \longleftarrow \texttt{Cohorting Algorithm}({\mathcal{V}}$)\\ 
            ${\boldsymbol \Theta}^{j} \longleftarrow {\boldsymbol \Theta}, \forall j \in [1,\abs{C}]$\\
            \BlankLine
            \textbf{Client Executes:} \\
            \texttt{ClientUpdates}(${\boldsymbol \Theta},k$) \\
        }
        \ElseIf{$r \geq 1$}{ 
            \textbf{Server Executes:} \\
            \For{$j=1\dots |\boldsymbol{C}| $}{
                \For{$k \in {C}^j$}{
                    ${\boldsymbol \Theta}^{j}_k \longleftarrow \texttt{ClientUpdates}({\boldsymbol \Theta}^{j},k) $
                }
                $ {\mathcal{V}^{j}} \longleftarrow \{\Theta^{j}_{k} | k \in C^{j}\} $ \\
                $ {\boldsymbol \Theta}^{j} \longleftarrow \operatorname{A}({\mathcal{V}^{j}}) $
            }
            Send ${\boldsymbol \Theta}^{j}$ to clients of $C^{j}$\\
            \textbf{Client Executes:} \\
            \texttt{ClientUpdates}(${\boldsymbol \Theta}^{j},k$) \\

        }
    }
    \BlankLine
    \BlankLine
    \SetKwFunction{FMain}{ClientUpdates}
    \SetKwProg{Fn}{}{:}{end}
    \Fn{\FMain{$\Theta$, $k$}}{
        $ (x_s, y_s) \forall s \in S_k $ \\
        Compute $\nabla \mathcal{L}_{k}(\boldsymbol{\Theta}) = \frac{1}{|S_k|} \sum_{s \in S_k} \nabla l_{s}(\boldsymbol{\Theta})$ \\
        $\boldsymbol{\Theta}_k \longleftarrow \boldsymbol{\Theta} - \eta\nabla \mathcal{L}_{k}(\boldsymbol{\Theta})$ \\
        Return ${\boldsymbol \Theta}_k$ to the Server
       }
    
   \caption{Lightweight Industrial Cohorted FL ($\texttt{LICFL}$)}
    \label{alg:cohort}
\end{algorithm}

In the next communication round ($r \geq 1$), the Algorithm~\ref{alg:cohort} updates the model parameters cohort-wise for building personalized models for the clients. For example, for the cohort $C^{j}$ the server sends the model $\boldsymbol \Theta^{j}$ to all the clients that belong to $j^{th}$ cohort, the same is described in line: $19$ of the Algorithm~\ref{alg:cohort}. Model updates from all the clients for $j^{th}$ cohort aggregated using an aggregation strategy $\operatorname{A}$ and the set of locally trained models ${\mathcal{V}}^{j}$ (line: $22$), where $\mathcal{V}^{j}$ represents set of locally updated model parameters from the clients that belong to the $j^{th}$ cohort ${\boldsymbol C}^{j}$. For our implementation of $\texttt{LICFL}$, we considered FedAvg \cite{a6} as aggregation strategy $\operatorname{A}$, but not limited to FedAvg, the Algorithm~\ref{alg:cohort} is agnostic to aggregation strategy. The aggregated models ${\boldsymbol \Theta}^j$ for $j \in [1,|\mathbf{C}|]$ are sent to the clients that belong to the cohort $C^{j}$ (line: $24$). Then client $j$ updates the local model (line: $26$) by using the received aggregated global model parameters ${\boldsymbol \Theta^{j}}$. This process repeats for every communication round $r \geq 1$ inside each cohort $C^{j}$.

\begin{algorithm}
    \SetAlgoLined
    \SetKwFunction{FMain}{Cohorting Algorithm}
    \SetKwProg{Fn}{}{:}{end}
    \Fn{\FMain{$\mathcal{V}$, $n$, $q$}}{
    \For{$k=0 \dots K$}{
        $\boldsymbol{\vx}_i \longleftarrow \texttt{Flatten}({\mathcal{V}}_k)$ \\
    }
    $\mX \longleftarrow [{\vx}^{T}_1, {\vx}^{T}_2 \dots {\vx}^{T}_K]^{T}$ \\
     \BlankLine
     $\boldsymbol{\lambda}, \mU \longleftarrow \textit{eig}({\mX}_{n}^{T}{\mX}_{n})$,where ${\mX_{n}}_{ij}=\mX_{ij}/(\sum_{j}\mX_{ij})^{1/2}$,$\forall (i,j)$ \\
     \BlankLine
     $\va,\mZ\longleftarrow$ Largest $n$ Eigen pairs $(\{\boldsymbol{\lambda}_{i}\}_{i=0}^{n}, \{\mU_{i}\}_{i=0}^{n})$ \\
     $\mY \longleftarrow \mX\mZ$\\
    $\mA_{ij} \longleftarrow \exp{\{-\|\vy_{i}^{T}-\vy_{j}^{T}\|/2\sigma^{2}\}}, \mA_{ii}=0, \forall (i,j)$ \\
    $\mD_{ii} \longleftarrow \sum_{j} \mA_{ij}, \mD_{ij}=0 (i\neq j),  \forall (i,j)$ \\
    $\mL\longleftarrow\mD^{-1/2} \mA \mD^{-1/2}$\\
    $\boldsymbol{\lambda}, \mU \longleftarrow \textit{eig}(\mL)$ \\
    $\vb,\mS\longleftarrow$ Largest $q$ Eigen pairs $(\{\boldsymbol{\lambda}_{i}\}_{i=0}^{q}, \{\mU_{i}\}_{i=0}^{q})$ \\
    $\mP_{ij}\longleftarrow \frac{\mS_{ij}}{(\sum_{j}\mS_{ij})^{1/2}},\forall (i,j)$  \\
    $C \longleftarrow \texttt{Clustering} (\{\boldsymbol{p}_{1}^{T},\boldsymbol{p}_{2}^{T},\cdots,\boldsymbol{p}_{K}^{T}\}) $\\
    return $C$ 
    }
    \caption{Model parameter-based cohorting}
    \label{alg:ch}
\end{algorithm}

We observed that different aggregation strategies perform well at different communication rounds (refer the Fig.\ref{fig3} for more insight). To leverage it for further improving global model convergence and performance, we propose an adaptive aggregation strategy selection method. The pseudo-code of the proposed adaptive aggregation selection algorithm is presented in Algorithm~\ref{alg:FL4}. The following section describes the same, which is a plug-and-play module (Algorithm~\ref{alg:FL4}) that improves global model convergence and the adaptability of federated learning inside each cohort by using the best suitable aggregation strategy. The algorithm uses only model parameters, without requiring any additional communications or computations at clients. The combination of $\texttt{LICFL}$ and the Algorithm~\ref{alg:FL4} is named as adaptive $\texttt{LICFL}~(\texttt{ALICFL}$).

\subsection{Auto selection of aggregation strategy}
\label{adaptive_aggregation}
Popular momentum-based global model optimization approaches such as FedAdam, FedAdaGrad, FedYogi \cite{b5} and QFedAvg \cite{b6}, propose aggregation strategies that account for the heterogeneity of the data among the clients by providing adaptability. However, they consider a fixed aggregation strategy for all communication rounds, which may not yield optimal performance for all communication rounds. The proposed Algorithm~\ref{alg:FL4}, leverages the advantages of different aggregation strategies (such as FedAdam, FedAdaGrad, and FedYogi) by choosing the best-suited one for the current communication round that improves the performance and speed of the model convergence.   

\begin{algorithm}[h]
    \SetAlgoLined
    \SetKwFunction{FMain}{A}
    \SetKwProg{Fn}{}{:}{end}
    \Fn{\FMain{${\mathcal{V}}^j$, $r$, ${\boldsymbol \Theta}$}}{
    Initialize: $\beta_{1},\beta_{2} \in (0,1)$ and $\tau,m_{0},v_{0}, {\boldsymbol \Theta}^{j}_{0}=0 $ \\
    ${\boldsymbol \Theta}^{j}_{r} = {\boldsymbol \Theta} $\\
    $\mathcal{V}^{j^{'}}_{k}=\mathcal{V}^{j}- {\boldsymbol \Theta}^{j}_{r}$   $\forall k $\\
   ${\Delta}^{j}_{r}= \sum_k \mathcal{V}^{j^{'}}_{k}/\abs{\mathcal{V}^{j^{'}}}$\\
    $m_{r}=\beta_{1}m_{r-1}+(1-\beta_{1}){\Delta}^{j}_{r}$\\
    $m_{r},v_{r} = 0 $ (\textbf{$\texttt{FedAvg}$}) \\
    $v_{r}=v_{r-1}+{(\Delta}^{j}_{r})^{2}$ (\textbf{$\texttt{FedAdagrad}$})\\
    $v_{r}=v_{r-1}-(1-\beta_{2})({\Delta}^{j}_{r})^{2}\text{sign}(v_{r-1}-({\Delta}^{j}_{r})^{2})$ $(\textbf{\texttt{FedYogi}})$\\
    $v_{r}=\beta_{2}v_{r-1}+(1-\beta_{2})({\Delta}^{j}_{r})^{2}$ (\textbf{$\texttt{FedAdam}$})\\
    ${\boldsymbol \Theta}^{j}_{r}={\boldsymbol \Theta}^{j}_{r-1}+\eta\frac{m_{r}}{\sqrt{v_{r}}+\tau}$ \\
    \BlankLine
    $ \mathcal{R} \longleftarrow \{{\boldsymbol \Theta}^{j}_{r} \text{ for all strategies}\}$ \\
    \BlankLine
    $ \operatorname{s} = {||{\boldsymbol \Theta}^{j}_{r}||}_{F} - {||{\boldsymbol \Theta}^{j}_{r-1}||}_{F} $ \\
    \BlankLine
    $ \mathcal{Q} \longleftarrow \{\operatorname{s} \text{ for all strategies}\}$ \\
    
    \BlankLine
    ${\boldsymbol \Theta}^{j}_{r} = \text{Value at index of }\texttt{Minimum}(\mathcal{Q}) \text{ in } \mathcal{R}$ \\
    $ {\boldsymbol \Theta}^{j}_{r-1} = {\boldsymbol \Theta}^{j}_{r} $ \\
    Return $ {\boldsymbol \Theta}^{j} = {\boldsymbol \Theta}^{j}_{r} $ 
    }
    \caption{Adaptive Strategy Selection}
    \label{alg:FL4}
\end{algorithm}

Algorithm~\ref{alg:FL4} takes a set of model parameters from all the clients $\mathcal{V}^{j}$ at the current communication round $r$ and the initial aggregated model parameters (or parameters from the previous communication round) at the server ${\boldsymbol \Theta}$ as the input. Further, the momentum parameters are computed for different aggregation strategies (lines: $6$ to $10$). Using the momentum parameters, Algorithm~\ref{alg:FL4} updates the model parameters ${\boldsymbol \Theta}^{j}_{r}$ for the current communication round $r$ for all the considered aggregation strategies (line: $11$). The set of updated model parameters from all the aggregation strategies is represented as $R$ (line: $12$). Next, the set of distances, measured as the Frobenius norm difference of the previous model parameters to the current model parameters are computed (shown in line: $13$). The set of distances represented by $\mathcal{Q}$ (shown in line: $14$). Finally, the Algorithm~\ref{alg:FL4} returns the aggregated model parameters by using the most suited aggregation strategy for the current communication round (line: $16$). This Algorithm~\ref{alg:FL4} would be directly integrated into Algorithm~\ref{alg:cohort} by calling it instead of a fixed aggregation strategy at line: $22$. The integrated algorithm is named Adaptive Lightweight Cohorted FL ($\texttt{ALICFL}$). 

\begin{figure*}[h]
	\centering
		\includegraphics[width=1.0\textwidth]{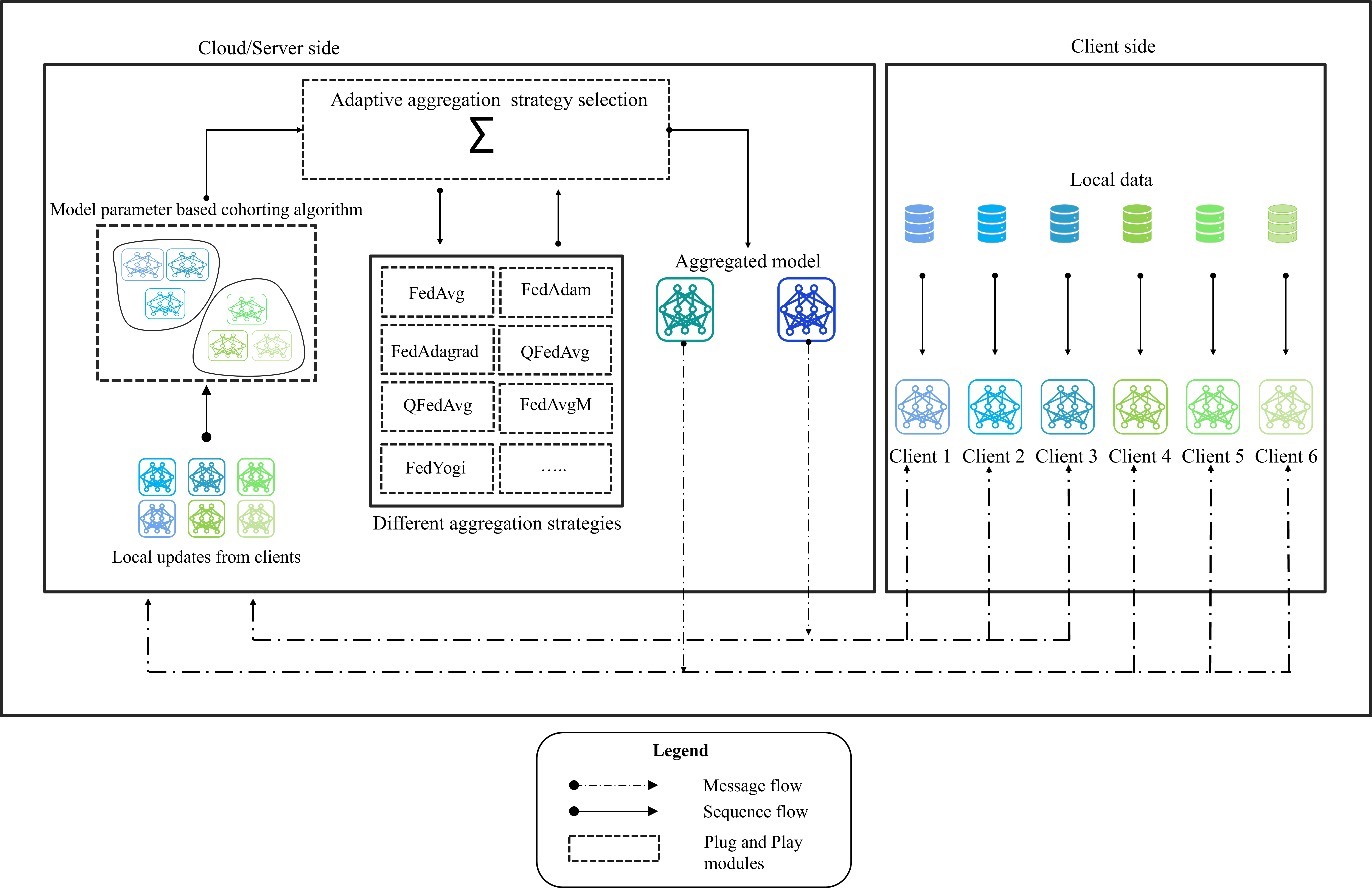}
        \caption{Functional architecture of Adaptive lightweight industrial cohorted federated learning ($\texttt{ALICFL}$) method.}
	\label{fig:aafl}
\end{figure*}

The functional architecture of the proposed $\texttt{ALICFL}$ method is shown in Fig.~\ref{fig:aafl}. The figure illustrates the modular structure of the proposed method. In the diagram "Message flow" refers to the exchange of model parameters between server and clients, and the "Sequence flow" refers to the data exchange within the server or a client. The architecture gives an overview of the proposed model parameter-based cohorting and adaptive strategy selection algorithms and illustrates their plug-and-play modular design, which provides the feasibility for seamless integration with any existing federated learning methods and facilitates on-the-fly adoption without the need for extensive modifications or overhauls. Fig.~\ref{fig:aafl} shows $6$ clients grouped into $2$ cohorts using the model parameter-based cohorting algorithm (Algorithm.~\ref{alg:ch}), models belonging to each cohort represented by a color variant. Two cohorts compute two different aggregated models using the adaptive aggregation strategy selection algorithm (Algorithm.~\ref{alg:FL4}) and share them back to the corresponding clients. 

\section{Experimentation}
In this section, we first describe the dataset used, followed by the ablation study of the neural network, which is built for the considered use case of predictive maintenance to demonstrate the performance of the proposed algorithms. Furthermore, the experimental setup is discussed.

\subsection{Data analysis}
We consider a use case of predictive maintenance where the objective is to forecast the instances of machine failures. We considered the predictive maintenance dataset \cite{b32}, a one-year time series comprising hourly data entries for 100 machines. The dataset includes measurements of voltage, rotation, pressure, and vibration. Additionally, it incorporates event logs and metadata such as each machine's size, make, model type, and age. It also contains the failure history of the machine and its components, maintenance records including proactive and reactive maintenance (component replacements), and error records detailing operational issues encountered by the machines (not categorized as failures). In the dataset, each machine is equipped with four components. The distribution of failures across these components is as follows: Comp1 accounts for 34.1\%, Comp2 for 25.2\%, Comp3 for 23.5\%, and Comp4 for 17.2\%. The dataset records 8761 entries per machine.   

We arranged the data set by mapping the instances of component failures along with the readings of each sensor, where $\vx_{i}=$[\textit{voltage, rotate, pressure, vibration}] is the considered input data at $i^{th}$ hour. The predictive maintenance is done based on the readings of the last $24$ hours for each machine. We consider a machine to have encountered failure $\vy_{i} =1$ if any of its components has an instance of failure in the past $24$ hours. In other words, $\vy_{i}=[Cp_{1} \cup Cp_{2} \cup Cp_{3}\cup Cp_{4}]$ where $Cp_{i}=1$ if there is a failure in $i^{th}$ component other wise it is $0$.

\subsection{Ablation study of predictive maintenance model}
We use hybrid neural networks that combine the strengths of two different neural networks. Since the data is a multivariate time series, we have employed an LSTM-CNN (Long Short Term Memory - Convolution Neural Network) hybrid network for the considered application. LSTMs extract temporal features, while CNNs are good in extracting spatial features. In our network, we have $2$ LSTM layers each of $100$ units with activation function \textit{tanh}, which are separated by RepeatVector layers. A single Dense layer with Linear activation is used at the output. In the CNN, we have a multi-headed CNN, $4$ CNN layers with a $2$ batch normalization layer. The CNN layers are stacked as follows: conv1 has $24$ filters with a kernel size of $4$, conv2 has $36$ filters of size $11$, conv3 has $48$ filters of size $3$ followed by batch normalization layer, conv4 has $32$ filters of size $3$ which is again followed by batch normalization layer. The \textit{Relu} activation is used for all CNN layers. After CNN layers, we have $3$ Dense layers stacked $(32, 16, 8)$ with \textit{Relu} activation. One Dense layer with \textit{Sigmoid} activation is used at the output layer. The data input will be in windowed form with a window size of $24$ with $4$ features. 

\subsection{Experimental setup}
To emulate the proposed Algorithm~\ref{alg:cohort}, we consider $100$ clients for our experimentation, where each machine ID from the data set is mapped as a client. The implementation is carried out using TensorFlow \cite{b11} and Flower framework \cite{b12,b7}. 

For evaluating the performance of proposed \texttt{LICFL} and $\texttt{ALICFL}$, we consider FL without Cohorting (labeled as \texttt{FL}), FL with cohorting using the statistical moments of the data \cite{a8} (labeled as $\texttt{IFL}$) as baseline approaches. We use $F_{1}$ score and mean squared error (loss) as our evaluation metrics.


\section{Results and Analysis}

\begin{figure}[htbp]
	\centering
		\includegraphics[width=0.5\textwidth]{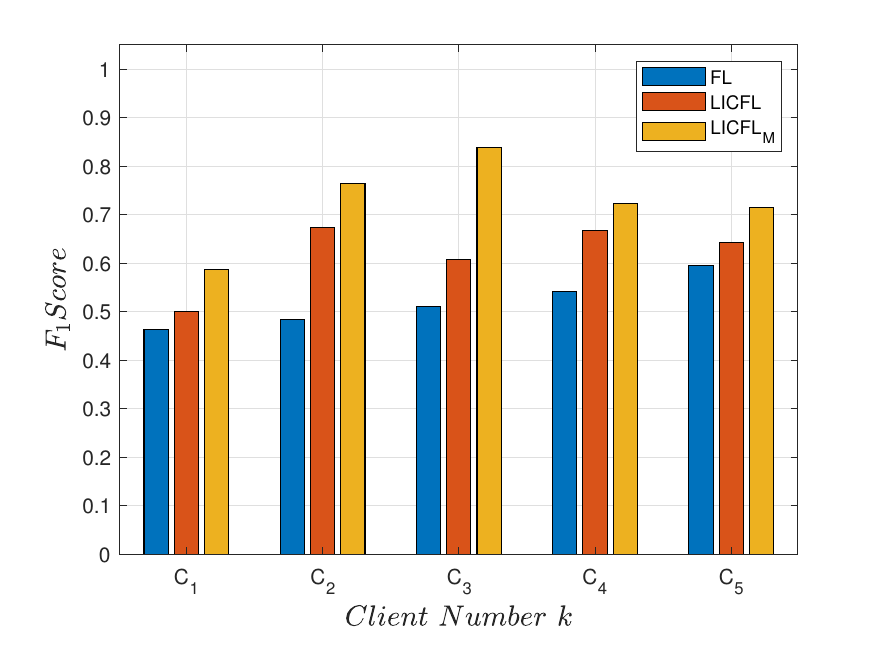}
        \caption{Effect of primary-level cohorting (using meta information) on client-level performance. Proposed algorithm $\texttt{LICFL}$ along with primary-level cohorting is labeled as $\texttt{LICFL}_{M}$. $\texttt{LICFL}_{M}$ outperforms the vanilla $\texttt{FL}$ and $\texttt{LICFL}$ at all considered clients.}
	\label{fig1}
\end{figure}

We conducted a set of experiments to demonstrate the effect of primary-level cohorting on model performance, where we used meta-information from the clients as filters or parameters for cohorting. The performance of the primary-level cohorting along with proposed $\texttt{LICFL}$ is shown in Fig.~\ref{fig1}. For demonstration, we considered client-level model performance after $30$ communication rounds (or after model convergence). In Fig.~\ref{fig1}, $\texttt{LICFL}_{M}$ denotes the combination of proposed $\texttt{LICFL}$ with primary-level cohorting using meta-information. Fig.~\ref{fig1} indicates that the cohorting based on filters (e.g. engine size, make year, and model number) that are derived from environmental and operational parameters (meta-information) improves the personalization of the client-model and hence, the client-level performance compared to vanilla $\texttt{FL}$ and $\texttt{LICFL}$ without considering primary-level cohorting. Note that for further experiments, we considered primary-level cohorting as default, proposed methodologies are applied along with it, and denoted $\texttt{LICFL}_{M}$ as $\texttt{LICFL}$ for easier representation.  

\begin{figure}[htbp]
	\centering
		\includegraphics[width=0.5\textwidth]{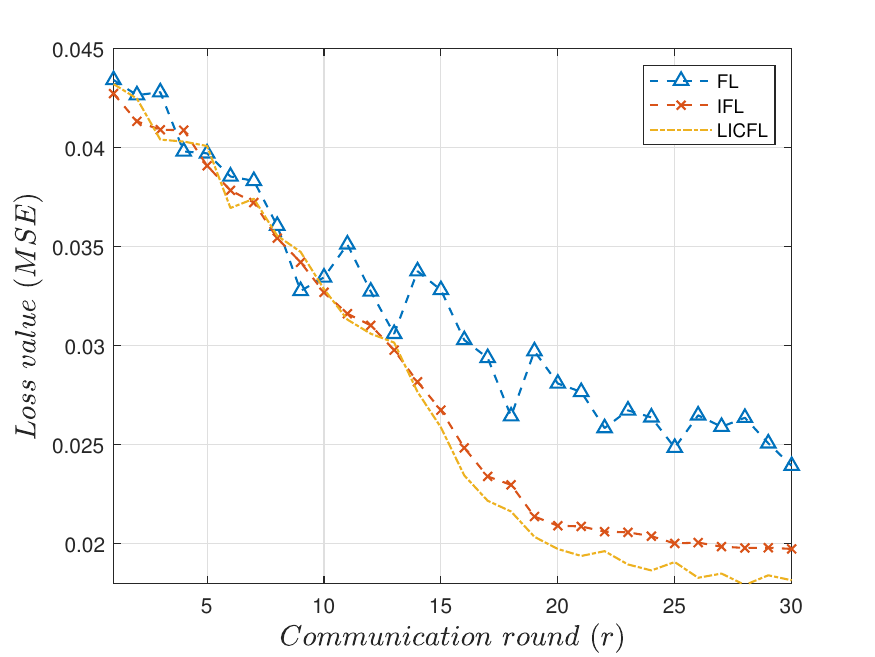}
        \caption{Comparison of global model performance at server against communication rounds. The proposed approach $\texttt{LICFL}$ provides a lower loss value and converges faster than baseline approaches.}
	\label{fig2}
\end{figure}

To evaluate the performance of the global model at the server against the communication rounds, we conducted experiments for $30$ communication rounds and the results are illustrated in Fig.~\ref{fig2}. Fig.~\ref{fig2}, shows the performance of the proposed method against recent and related approaches. Specifically, server model performance is measured as the Mean Square Error ($MSE$) value (loss) of the global consensus model with test data at the server for different communication rounds. From Fig.~\ref{fig2}, one can understand that an increase in communication rounds increases global model performance (decreases the loss value). It can be observed that the proposed approach provides lower loss values and converges faster than the baseline approaches.   

\begin{figure}[htbp]
	\centering
		\includegraphics[width=0.5\textwidth]{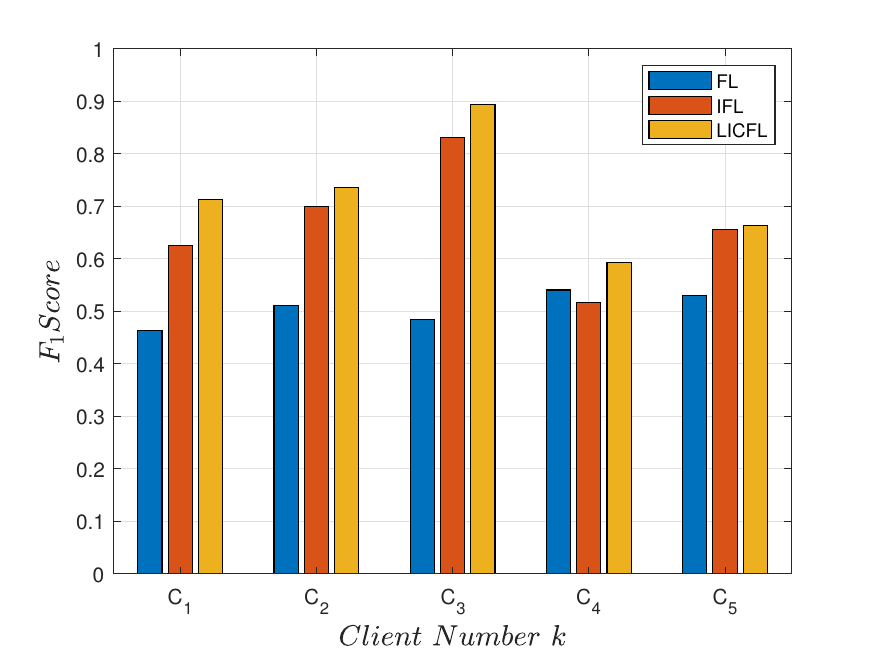}
        \caption{Comparison of client-level performance of the proposed method against considered baseline approaches. The proposed algorithm $\texttt{LICFL}$ demonstrates superior performance compared to considered baseline approaches for all randomly picked clients.}
	\label{fig3}
\end{figure}

To evaluate the client-level performance, we randomly pick $5$ clients from the considered $100$  to evaluate the client-level performance clients. Specifically, client-model or client-level performance is defined as the global model performance with on-client test data. Fig.~\ref{fig3}, shows the client-model performance of the proposed method against considered baseline approaches. One can observe that, the proposed approach outperforms the baseline approaches for all clients as shown in Fig.~\ref{fig3}.

\begin{figure}[htbp]
	\centering
		\includegraphics[width=0.5\textwidth]{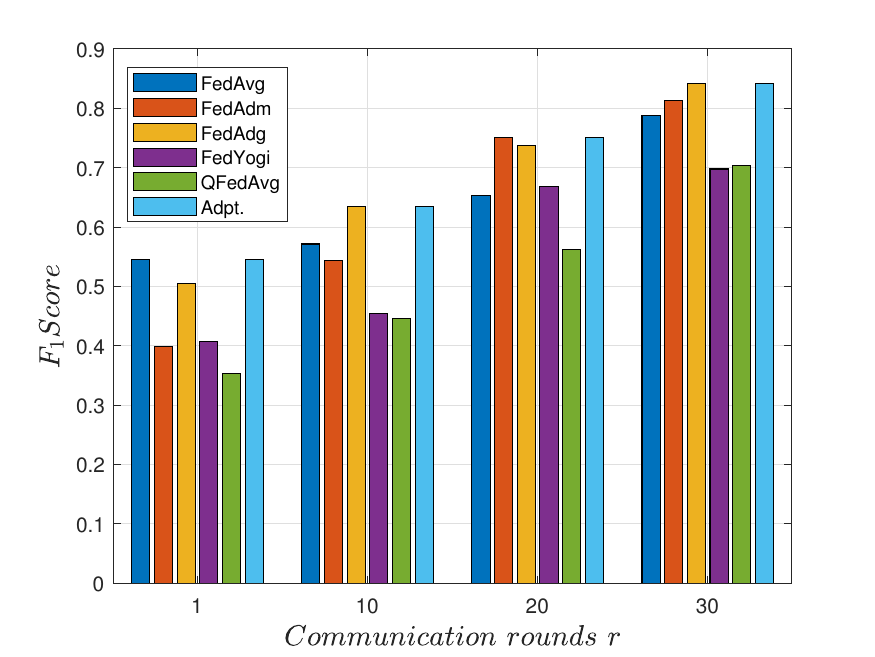}
        \caption{Comparison of a randomly chosen client-level performance of the proposed aggregation strategy $\texttt{Adpt}$. The results indicate that the $\texttt{Adpt}$ adapts to the best-performing aggregation strategy for every communication round and provides superior performance compared to others.}
	\label{fig4}
\end{figure}

Another set of experiments was conducted to illustrate the performance of the proposed adaptive aggregation strategy for $30$ communication rounds. For the evaluation, popular aggregation strategies such as $\texttt{FedAvg}$, $\texttt{FedAdam}$, $\texttt{FedAdagrad}$, $\texttt{FedYogi}$, $\texttt{QFedAvg}$ were considered as baseline approaches. For a fair comparison, we employed $\texttt{LICFL}$ as the baseline federated learning algorithm for evaluating all the considered aggregation strategies. Fig.~\ref{fig4} presents a randomly chosen client-level performance of the proposed adaptive aggregation strategy algorithm, labeled as $\texttt{Adpt}$ along with considered baseline aggregation strategies. One can observe that the proposed adaptive aggregation approach ($\texttt{Adpt}$) outperforms all aggregation strategies by dynamically switching to the most effective strategy at that particular communication round for aggregating the weight matrices from the clients. For example in the first communication round $r=1$, the proposed adaptive aggregation algorithm $\texttt{Adpt}$ switches to best performing \texttt{FedAvg}, similarly for $r=2$, $\texttt{Adpt}$ switches to $\texttt{FedAdg}$.

\begin{figure}[htbp]
	\centering
		\includegraphics[width=0.5\textwidth]{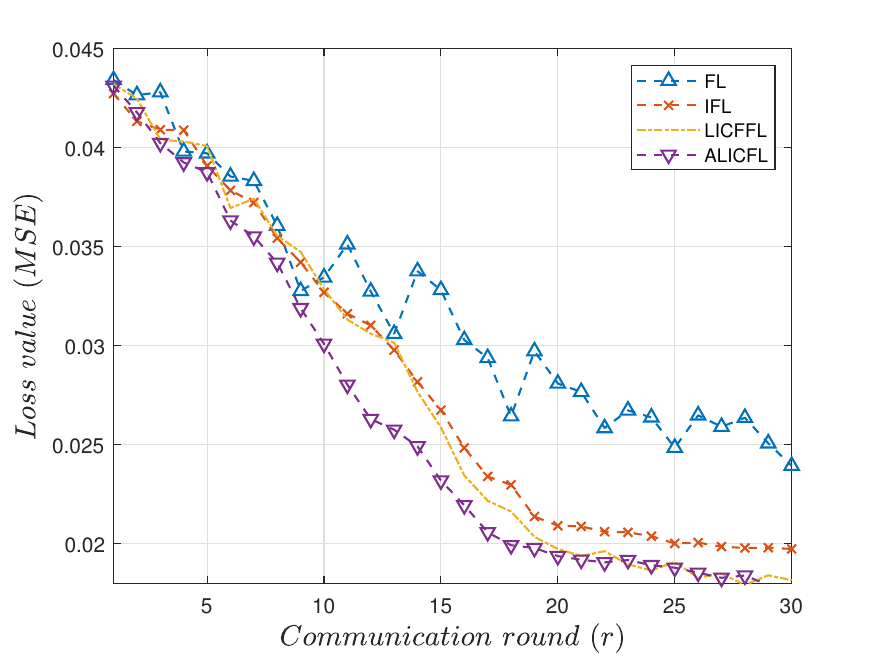}
        \caption{Comparison of global model performance of the proposed  $\texttt{ALICFL}$ against communication rounds and considered baseline approaches. The proposed approach provides superior performance (with lower loss value) and converges faster than considered baseline approaches.}
	\label{fig5}
\end{figure}

To illustrate the performance of the proposed adaptive aggregation strategy $\texttt{Adpt}$ in integration with the proposed methodology $\texttt{LICFL}$ against communication rounds, an experiment for $30$ communication rounds was conducted. We integrate $\texttt{Adpt}$ as a plug-and-play module to the proposed $\texttt{LICFL}$ methodology and the resulting method called $\texttt{ALICFL}$, detailed description is in  Section~\ref{adaptive_aggregation}. The experimental results are shown in Fig.~\ref{fig5}. From Fig.~\ref{fig5}, one can understand that the proposed $\texttt{ALICFL}$ algorithm consistently provides the best performance with lower loss values and converges faster than the considered baseline approaches for all communication rounds. 

\begin{figure}[htbp]
	\centering
		\includegraphics[width=0.5\textwidth]{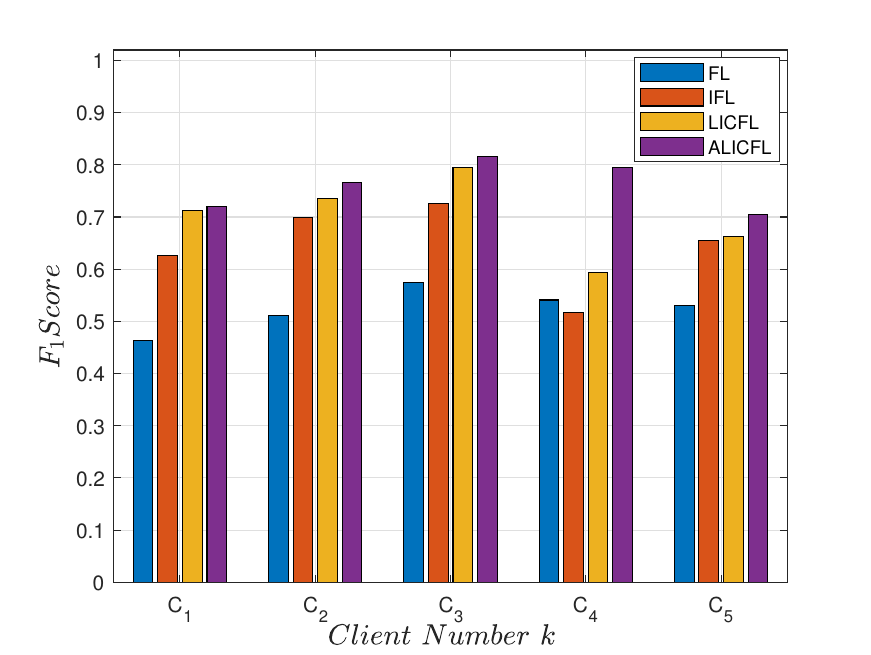}
        \caption{Comparison of client-level performance of the proposed method $\texttt{ALICFL}$ against considered baseline approaches. The proposed approaches outperform the considered baseline approaches for all randomly picked clients.}
	\label{fig6}
\end{figure}

As mentioned before, for evaluating the client-level performance of the proposed $\texttt{ALICFL}$, again $5$ clients were randomly selected among $100$. The results are illustrated in Fig.~\ref{fig6}. The results are considered after $30$ communication rounds or convergence. From Fig.~\ref{fig6}, one can observe that the proposed approach $\texttt{ALICFL}$ outperforms the considered baseline approaches for all the considered clients.

\section{Conclusion}
In this paper, we proposed a novel Lightweight Industrial cohorted FL ($\texttt{LICFL}$) algorithm that clusters similar clients to avoid adverse effects in the performance raising due to the presence of heterogeneous data distributions across the clients. Our algorithm uses model parameters to cohort the clients and does not need any additional on-edge computations and communications. Our approach enhances the client-level model performance by allowing them to train more specialized or personalized models. Also, proposed an algorithm for auto selection for aggregation strategies per cohort, which extends the $\texttt{LICFL}$ to Adaptive $\texttt{LICFL}$ ($\texttt{ALICFL}$) for further improving the client-level model performance and speeds up the global model convergence. Through the numerical experiments, we showed that the proposed approaches $\texttt{LICFL}$ and $\texttt{ALICFL}$ outperform the considered baseline approaches.

\end{document}